\documentclass[11pt,a4paper]{article}
\usepackage{times,latexsym}
\usepackage{url}
\usepackage[T1]{fontenc}
\usepackage{booktabs}
\usepackage{caption}
\usepackage{graphicx}
\usepackage{amsmath,amsfonts}
\usepackage{bm}
\usepackage{algorithm}
\usepackage{algpseudocode} 
\usepackage{multirow} 
\usepackage[table,xcdraw]{xcolor} 
\usepackage{subfig}
\usepackage{varwidth}
\usepackage{enumitem}

\newcommand{\red}[1]{\textcolor{red}{#1}}
\newcommand{\blue}[1]{\textcolor{blue}{#1}}

\usepackage[acceptedWithA]{tacl2021v1}
\usepackage{xspace,mfirstuc,tabulary}

\newif\iftaclinstructions
\taclinstructionsfalse % AUTHORS: do NOT set this to true
\taclinstructionstrue
\iftaclinstructions
\renewcommand{\confidential}{}
\renewcommand{\anonsubtext}{(No author info supplied here, for consistency with TACL-submission anonymization requirements)}
\newcommand{\instr}
\fi

\iftaclpubformat % this "if" is set by the choice of options

\else

\fi

\title{Metropolis-Hastings Captioning Game: Knowledge Fusion of Vision Language Models via Decentralized Bayesian Inference}

\author{
\textbf{Yuta Matsui}$^{1}$ \quad
\textbf{Ryosuke Yamaki}$^{1}$ \quad
\textbf{Ryo Ueda}$^{2}$ \quad \\
\textbf{Seitaro Shinagawa}$^{3}$ \quad
\textbf{Tadahiro Taniguchi}$^{4,1}$ \\
$^{1}$Ritsumeikan University \quad
$^{2}$The University of Tokyo \quad \\
$^{3}$Nara Institute of Science and Technology \quad
$^{4}$Kyoto University \\
\texttt{\{matsui.yuta, yamaki.ryosuke\}@em.ci.ritsumei.ac.jp}, \\
\texttt{ryoryoueda@is.s.u-tokyo.ac.jp}, \\
\texttt{sei.shinagawa@is.naist.jp},~
\texttt{taniguchi@i.kyoto-u.ac.jp}
}

\date{}

\begin{document}
\maketitle
\begin{abstract}
% We propose the Metropolis-Hastings Captioning Game (MHCG) for fusing pre-trained Vision-Language Models (VLMs). Existing methods that combine multiple models suffer from increased inference costs and architectural constraints. Our approach formalizes the process where VLM agents infer and learn from shared captions for images as decentralized Bayesian inference, enabling effective knowledge fusion.
% In experiments, applying MHCG to two VLM agents pre-trained on different datasets led to a maximum X\% improvement in captioning performance measured by CLIPScore on the partner’s pre-training data. We also constructed a split COCO dataset and evaluation metrics to isolate and evaluate category-specific knowledge. Even in challenging scenarios with agents possessing distinct category-specific knowledge, MHCG outperformed ensemble and weight-averaging methods, demonstrating superior category-level comparison in the captioning task.
We propose the Metropolis-Hastings Captioning Game (MHCG), a method to fuse knowledge of multiple vision-language models (VLMs) by learning from each other. Although existing methods that combine multiple models suffer from inference costs and architectural constraints, MHCG avoids these problems by performing decentralized Bayesian inference through a process resembling a language game. The knowledge fusion process establishes communication between two VLM agents alternately captioning images and learning from each other.
We conduct two image-captioning experiments with two VLMs, each pre-trained on a different dataset. The first experiment demonstrates that MHCG achieves consistent improvement in reference-free evaluation metrics. The second experiment investigates how MHCG contributes to sharing VLMs' category-level vocabulary by observing the occurrence of the vocabulary in the generated captions.
\end{abstract}

\section{Introduction}
There has been growing interest in methods that efficiently combine the capabilities of multiple pre-trained VLMs, each possessing different knowledge, through their fusion.
Large-scale VLMs for vision-language tasks, such as CLIP and BLIP \citep{radford2021learning, li2022blip}, have been pre-trained based on diverse architectures and datasets.
It is well known that the datasets of image-text pairs used for this pre-training are relatively small in scale compared to text corpora, due to the high cost of data collection and annotation \citep{srinivasan2021wit, alayrac2022flamingo}.
Therefore, developing methods to effectively leverage the diverse pre-trained models with different knowledge by fusing them is of great importance.

% 2. Research on model integration
% There have been numerous attempts to develop methods that effectively combine the knowledge of each model by integrating multiple pre-trained models.
% To compensate for the limitations of individual models, ensemble learning and weight averaging have been employed in fields such as image recognition and natural language processing \citep{li2023deep}.
Numerous methods have been attempted to fuse multiple pre-trained VLMs, such as ensemble learning and weight averaging in the fields of image recognition and natural language processing \citep{li2023deep}.
Although ensemble learning combines the outputs of each model, it increases the required computational resources and time cost during inference, as all models need to be executed \citep{liu2021dexperts, mavromatis2024pack}.
Weight averaging, which directly merges the parameters, requires the models to share the same architecture, and there are concerns about performance degradation when merging models trained with different initializations or datasets \citep{wortsman2022modelsoup, yang2024adamerging}.

\begin{figure*}[tb!]
    \centering
    \subfloat[Graphical model of the probabilistic generative model ]{%
        \raisebox{1.2cm}{%
        \includegraphics[width=0.25\textwidth,keepaspectratio]{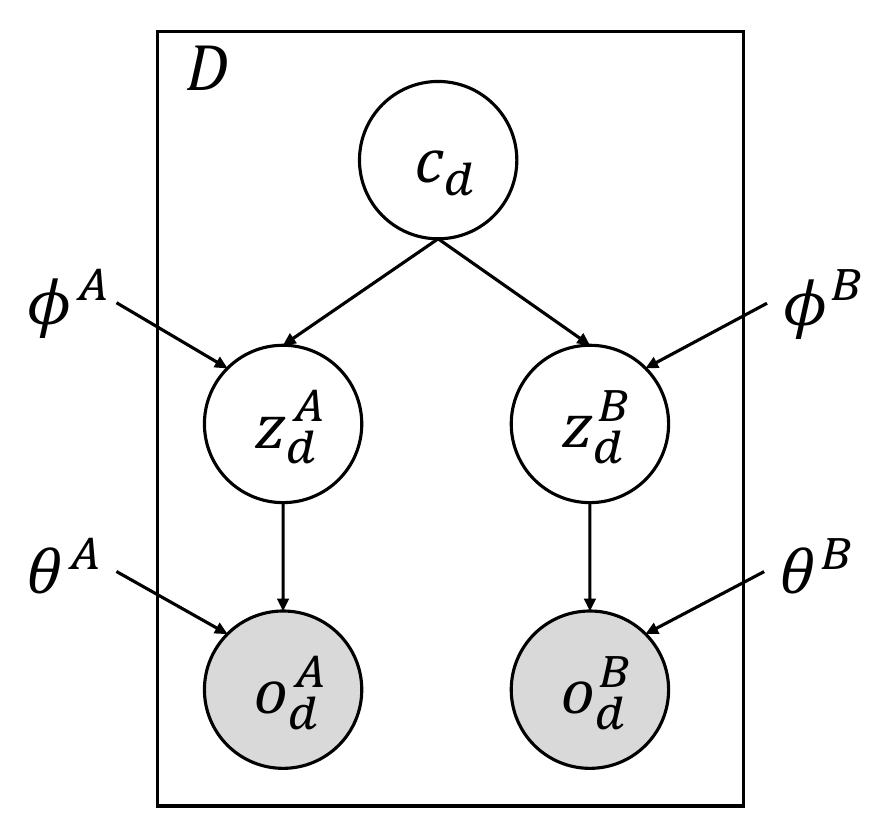}%
        }%
        \label{fig:graphical_model}%
    }%
    \hspace{0.01\textwidth}
    \subfloat[Estimation of intractable posterior \( p(c | o^{A},o^{B};\Theta^{A},\Theta^{B}) \)]{%
        \includegraphics[width=0.3\textwidth,keepaspectratio]{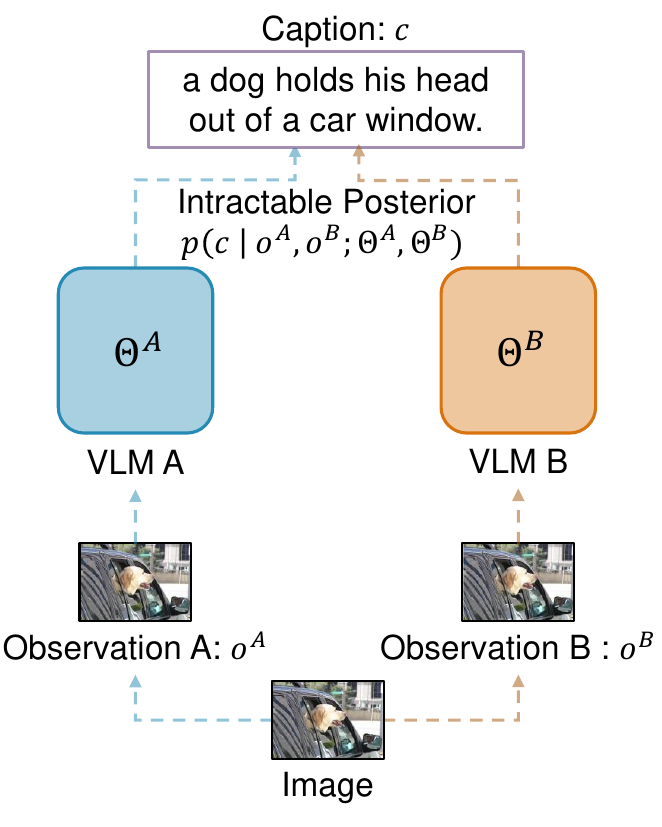}%
        \label{fig:overview_of_research}%
    }%
    \hspace{0.01\textwidth}
    \subfloat[Overview of the Metropolis Hastings Captioning Game]{%
        \includegraphics[width=0.4\textwidth,keepaspectratio]{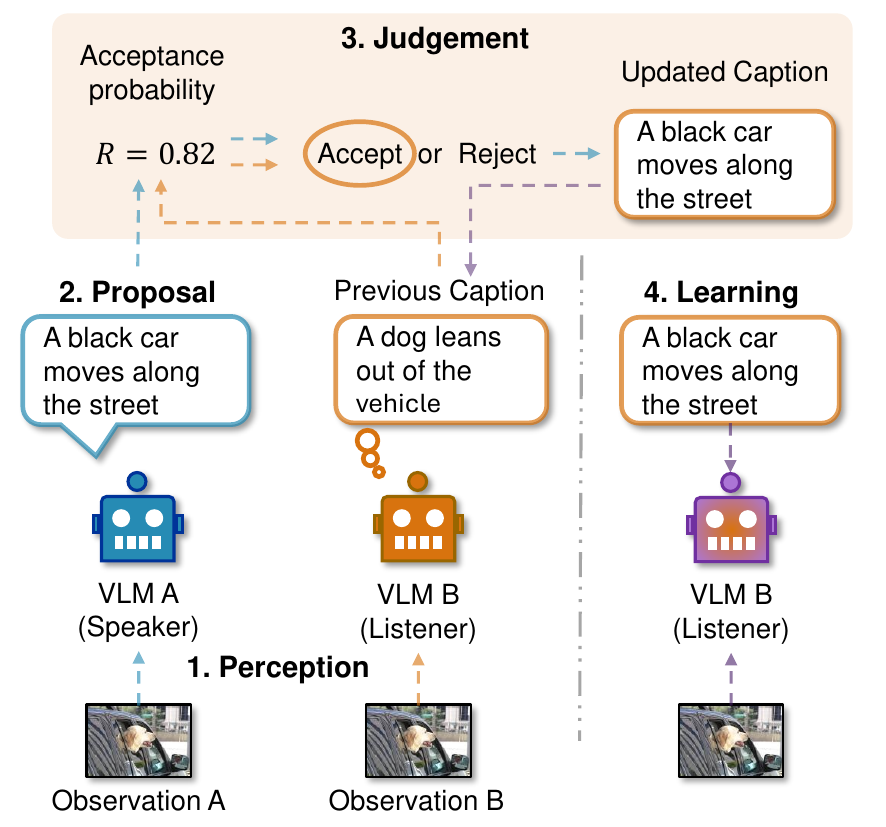}%
        \label{fig:overview_of_mhcg}%
    }%
    \caption{Overview of this research.}
    \label{fig:combined_figures}
\end{figure*}

% ここに何かしらの方向転換が欲しい気がする(Ueda)
To alleviate the limitations of the previous works, we reinterpret the VLM fusion problem from the Bayesian perspective.
Here, we give a rough sketch of our formulation (a full description will be given in Section \ref{sec:method}).
Suppose that we have two pre-trained VLMs \(p(c | o^{A};\Theta^{A})\) and \(p(c | o^{B}; \Theta^{B})\) parametrized by \(\Theta^{A}\) and \(\Theta^{B}\) respectively, where \(c\) is a caption and \(o^{A},o^{B}\) are observations (i.e., images).
A ``hidden'' generative model \( p(o^{A}, o^{B} | c; \Theta^{A},\Theta^{B}) \) being assumed as shown in Figure~\ref{fig:graphical_model}, the VLM fusion boils down to the estimation problem of intractable posterior \( p(c | o^{A},o^{B};\Theta^{A},\Theta^{B}) \) (as presented in Figure~\ref{fig:overview_of_research}).
One might notice at this point that the previous works approximated the posterior only in an ad-hoc manner, e.g., \( p(c | o^{A}; \Theta^{A})p(c | o^{B}; \Theta^{B})\) in ensemble learning and \( p(c | o^{A}; (\Theta^{A}+\Theta^{B})/2) \) in weight averaging, which might cause a failure of diverse knowledge fusion.
In contrast, we pursue a more accurate estimation via Markov chain Monte-Carlo (MCMC) and adopt an EM-like learning algorithm to avoid inefficiency in inference as seen in ensemble methods.

% この辺でバシッと提案手法を言うべきな気がする(Ueda)
% "5. Our Research"のフレーズをいっぱい借りつつ、ここで提案手法を記述してみる(Ueda)
To this end, we propose the Metropolis-Hastings Captioning Game (MHCG) that fuses the diverse knowledge held by multiple pre-trained VLM agents.
In this study, VLM agents are represented as Inter-ProbVLMs within a probabilistic generative model (PGM) and adopt an approach inspired by the Metropolis-Hastings Naming Game \citep[MHNG,][]{taniguchi2023emergent}, as illustrated in Figure~\ref{fig:overview_of_mhcg}.
The agents probabilistically accept or reject proposed captions based on the Metropolis-Hastings algorithm, thereby approximating the most plausible caption for both.
By learning from the inferred captions, we aim to achieve effective knowledge fusion.

% 3と4をうまくマージしつつ簡潔な雰囲気にしたい。
% ついでにいろんなEmCom論文をここで引用することで、”鎮め”たい
Our approach with MHCG is largely inspired by the MHNG in Emergent Communication (EmCom), a research field that studies how communication protocols emerge among artificial agents through interactions in a game.
While early EmCom studies can be traced back a few decades \citep[e.g.,][]{Briscoe2000grammatical, Kirby2001spontaneous, Steels2015talking, Spranger2012perceptual}, it has recently gained renewed interest due to the development of deep reinforcement/representation learning \citep{lazaridou2020emergent, boldt2024review, Rita2024language}.
Among such, a notable aspect of MHNG is that some sort of communication can be seen as MCMC, namely the Metropolis-Hastings (MH) sampling algorithm.
In other words, communication among agents can be formulated as a posterior estimation via MCMC.
This makes our MHCG formulation interesting and appealing as it allows us to view the VLM fusion problem as an intuitive communication process between agents; otherwise, the problem would remain just pedantic and less intuitive.

Our main contributions are proposing MHCG (Section~\ref{sec:method}) and revealing its effectiveness through two experiments. 
The first experiment compares VLMs pre-trained by different datasets (COCO, CC3M) and demonstrates that MHCG contributes to achieve consistent improvement in reference-free evaluation metrics (Section~\ref{sec:experiment1}). 
The second experiment provides a more detailed analysis of how MHCG works through a vocabulary-level comparison using COCO categories (Section~\ref{sec:experiment2}).

% \begin{figure}[tb] 
% \centering \includegraphics[width=0.47\textwidth]{images/overview_of_MHCG.pdf} \caption{Overview of the Metropolis Hastings Captioning Game.} \label{fig:overview of mhcg} 
% \end{figure}

\fi

\section{Related Work}
\paragraph{Model Fusion} \label{sec:weight_averaging}
% Ensemble learning~\cite{mienye2022ensemble_survey} is an approach that achieves high overall accuracy by ensembling the outputs of multiple models. Methods include simply averaging the outputs, computing the difference between expert and anti-expert models, and weighting by perplexity. 
% This technique effectively leverages the complementary strengths of each model’s predictions. However, it imposes constraints—such as reduced inference speed—since it requires multiple models to operate simultaneously during inference.
Model fusion methods~\cite{gao-etal-2022-revisiting,huang2017snapshot,matena2022merging,Wang2020Federated} have been widely-explored as a counterpart of ensemble learning~\cite{mienye2022ensemble_survey}, which achieves high performance by leveraging multiple model outputs in a sequential or parallel inference algorithm.
These methods have an advantage at reduction of inference cost by unifying model parameters.

A representative model fusion is averaging checkpoints such as model averaging~\cite{gao-etal-2022-revisiting} and Snapshot Ensembling~\cite{huang2017snapshot}.
Another averaging method determines the parameters based on the Fisher information matrix~\cite{matena2022merging}.
Recent interest of model fusion extends to model merging~\cite{pmlr-v162-wortsman22a,du-etal-2024-knowledge,akiba2025evolutionary}, a field to add new capability of a task by considering arithmetic between models.
However, these methods work in the limited situation that the target models to be fused are similar enough.
There exists a risk of significant performance degradation after fusion if the models are distinct due to differences in the training's initial conditions or pre-training data.
Despite the limitations, MHCG does not require consideration of weighting methods for fusing and works between the distinct models.
% Weight averaging is a technique that directly merges the parameters of multiple pre-trained models. Some approaches simply average the parameters, while others determine the weights based on the Fisher information matrix~\cite{matena2022merging}. Although the computational efficiency of this method is appealing, it requires that the parameters of each model have the same shape for successful fusion. Additionally, there exists a risk of significant performance degradation after fusion due to differences in initial conditions or training data.

\paragraph{Emergent Communication}
The development of deep reinforcement and representation learning has enabled neural agents to give rise to communication protocols from scratch (i.e., without any natural language supervision) by learning to communicate \citep{foerster2016learning, lazaridou2017multiagent, havrylov2017emergence}, which revived the research direction called Emergent Communication (EmCom).
On the one hand, most existing EmCom studies are in the field of computational linguistics, i.e., they typically investigated whether the emerged protocols exhibited similar statistical properties as human language, in the settings of language emergence from scratch \citep{kottur2017natural, chaabouni2019antiefficient, chaabouni2020compo, ueda2023word}.
On the other hand, importing communication game frameworks from the EmCom field, recent studies utilize them as fine-tuning methods of pre-trained language models \citep{baroni2022emergent, mahaut2023referential, dessi2023cross}.
We take the latter direction.
Specifically, our study is largely inspired by the Metropolis-Hastings Naming Game (MHNG), which models the process of symbol emergence as decentralized Bayesian inference within a probabilistic generative model (PGM) \citep{taniguchi2023emergent}, based on the collective predictive coding (CPC) hypothesis~\cite{taniguchi2024collective}.
% In MHNG, agents collaboratively assign names to observed objects through iterative proposals and probabilistic acceptance or rejection, eventually reaching a consensus on shared names.
% This process is formulated as posterior inference using the Metropolis-Hastings algorithm, allowing agents to perform inference independently without relying on each other's internal states \cite{taniguchi2024collective}.
% Previous studies have introduced methods in which agents share categorical names \cite{} and approaches that address linguistic compositionality by considering combinations of words \cite{}.

% This study extends MHNG by introducing vision-language model (VLM) agents equipped with prior linguistic knowledge. The framework enables agents to share captions generated from image observations as signs, collaboratively infer captions based on their prior knowledge, and learn from this process. As a result, the proposed approach aims to effectively integrate the knowledge of both agents through cooperative learning.

\section{Method}
\label{sec:method}

To realize a communication-based knowledge fusion as motivated in the previous section, we extend MHNG \citep{taniguchi2023emergent} to formulate the \emph{Metropolis-Hastings Captioning Game} (MHCG), which will be described in Section \ref{sec:metropolis-hastings-captioning-game}.
MHNG models a communication process of giving names to observed objects (N for ``Naming''), while MHCG models a similar process of giving captions to given images (C for ``Captioning'').
Thus, we need to replace probabilistic models of fixed dimensional vectors (names) with those of variable-length discrete sequences (captions).
We will handle this in Section \ref{sec:inter-probvlm} using what we dub \emph{Inter-ProbVLM}.

\subsection{Metropolis-Hastings Captioning Game}
\label{sec:metropolis-hastings-captioning-game}

% The Metropolis-Hastings Captioning Game (MHCG) is an extension of MHNG \citep{taniguchi2023emergent}, which assigns categorical labels (i.e., naming) to objects, by applying it to the generation of captions (i.e., captioning) for images. 
In MHCG, two pre-trained agents, $A$ and $B$, communicate via mutual proposals and acceptance/rejection decisions in order to fuse their respective information. 
At each round, one of them is assigned as a \emph{speaker}, who proposes a caption based on a given image.
The other is assigned as a \emph{listener}, who decides whether to accept the proposed caption and updates its parameter accordingly upon acceptance.
Their roles are flipped at the next round.
By iterating the rounds, the two agents are expected to converge to an agreement on the most plausible caption for both. 
Specifically, MHCG comprises the following four steps: Perception, Proposal, Judgment, and Learning as explained follows.

\paragraph{Perception}

We denote by $*\in\{A, B\}$ the index of an agent.
Both agents obtain a latent representation $z_d^*$ from the $d$-th observation $o_d^*$.  
This process is realized by sampling from the image encoder parameterized by $\xi$, as shown in Equation (\ref{eq:perception}). The latent representation $z_d^*$ reflects each agent's perception of the observation and serves as the basis for proposing captions and acceptance decisions.
\begin{equation}
    z^{*}_d \sim q(z^{*}_d|o^{*}_d;\psi^{*}).\label{eq:perception}
\end{equation}
\paragraph{Proposal}

The speaker agent proposes a caption $c_d^{\star}$ based on the latent representation $z_d^{Sp}$.
This is achieved by sampling from the text decoder parameterized by $\xi$, as shown in Equation (\ref{eq:propose}).
\begin{equation}
    c^{\star}_d \sim q(c|z^{Sp}_d;\xi^{Sp})\label{eq:propose}.
\end{equation}
\paragraph{Judgment}

The listener agent judges whether to accept the proposed sign $c_d^{\star}$ based on its own perception $z_d^{Li}$ and its internally recognized sign $c_d^{Li}$.
Notably, this corresponds to an MCMC-based sampling algorithm, namely the Metropolis-Hastings (MH) algorithm, from the posterior distribution $p(c_d | z^A_d, z^B_d, \phi^A, \phi^B)$, where the proposal distribution is defined by Equation (\ref{eq:propose}).
The listener agent accepts $c_d^{\star}$ with acceptance probability $r = \min(1, R)$, where $R$ is given by:
\begin{align}
    R &= \frac{p(c^{\star}_d \mid z^{Sp}_d, z^{Li}_d; \phi^{Sp}, \phi^{Li})q(c^{Li}_d \mid z^{Sp}_d;\xi^{Sp})}{p(c^{Li}_d \mid z^{Sp}_d, z^{Li}_d; \phi^{Sp}, \phi^{Li})q(c^{\star}_d \mid z^{Sp}_d;\xi^{Sp})} \notag \\
   %  &=\frac{p(z^{Li} \mid c^{\star};\phi^{Li})p(c^{\star}\mid z^{Sp}; \phi^{Sp})q(c^{Li}\mid z^{Sp}; \xi^{Sp})}
   % {p(z^{Li} \mid c^{Li};\phi^{Li})p(c^{Li}\mid z^{Sp}; \phi^{Sp})q(c^{\star}\mid z^{Sp}; \xi^{Sp})} \notag\\
   &\approx 
    \frac{
    p(z^{Li}_d \mid c^{\star}_d; {\phi}^{Li})}
    {
    p(z^{Li}_d \mid c^{Li}_d;{\phi}^{Li})
    }.\label{eq:acceptance probability}
\end{align}
Here, we introduce an approximation, replacing the intractable true posterior $p(c^{Sp}_d \mid z^{Sp}_d; \phi^{Sp})$ with the approximate posterior used in the proposal distribution $q(c^{Sp}_d \mid z^{Sp}_d; \xi^{Sp})$ \citep{hoang2024compositionality}.
% This approximation is valid when the text encoder $\phi^{*}$ and text decoder $\xi^*$ are well-trained on $c^*$ and $z^*$.
% The proposal and judgment steps together constitute the ``Metropolis-Hastings Communication'' process, as shown in Algorithm \ref{al:MH-Communication}.

\paragraph{Learning}

The listener agent updates its parameters $\xi^{Li}, \phi^{Li}, \psi^{Li}, \theta^{Li}$ based on the latent representation $z^{Li}_d$ and the accepted caption $c^{Li}_d$.
The agent incrementally adapts its parameters to better align with the captions proposed by the other agent in subsequent iterations.
\begin{align}
    L_{\xi} = & -\mathbb{E}_{p(c \mid z^{Li},\,z^{Sp};\,\phi^{Li},\,\phi^{Sp})}
    \Big[ \log q(c^{*}\!\mid\!z^{*};\,\xi^{*}) \Big] \label{eq:loss for xi} \\
    L_{\phi} = & -\mathbb{E}_{p(c\mid z^{Li},\,z^{Sp};\,\phi^{Li},\,\phi^{Sp})}
    \Big[ \log p(z^{*}\!\mid\!c^{*};\,\phi^{*}) \Big] \label{eq:loss for phi} \\
    L_{\psi} = & -\mathbb{E}_{p(z^* \mid c^{*};\,\phi^{*})}
    \Big[ \log q(z^{*} \mid o^{*};\,\psi^{*}) \Big] \label{eq:loss for psi} \\
    L_{\theta} = & -\mathbb{E}_{p(z^{*} \mid c^{*};\,\phi^{*})}
    \Big[ \log p(o^{*} \mid z^{*};\,\theta^{*}) \Big] \label{eq:loss for theta}
\end{align}
To prevent catastrophic forgetting \citep{mccloskey1989catastrophic}, we introduce Low Rank Adaptation \citep[LoRA,][]{hu2021lora} and Dark Experience Replay++ \citep[DER++,][]{buzzega2020dark} during parameter updates.
DER++ is a continual learning method that maintains past task logits and ground-truth captions in a replay buffer, enabling adaptation to sudden distribution shifts.
This ensures that each agent retains its prior knowledge while adapting to the captions proposed by the other agent.
DER++ is a method designed to mitigate forgetting by incorporating an additional term that aligns with previous model outputs:
\begin{align}
L_{\xi} = & -\mathbb{E}_{p(c \mid z^{Li}, z^{Sp}; \phi^{Li}, \phi^{Sp})}
\Big[ \log q(c \mid z^{*}; \xi^{*}) \Big] \notag \\
 +& \alpha \, \mathbb{E}_{(z', c', h') \sim M^{*}_{\xi}} 
\Big[ \| h' - h_{\xi^{*}}(z') \|_2^2 \Big] \notag \\
 -& \beta \, \mathbb{E}_{(z'', c'', h'') \sim M^{*}_{\xi}} 
\Big[ \log q(c'' \mid z''; \xi^{*}) \Big] \label{eq:loss for xi with DER++}
\end{align}
Here, $M^{*}_{\xi}$ represents a buffer initialized using each agent's pre-training data. 
The sampled variables $z', z''$ correspond to latent representations extracted from images in the pre-training dataset, while $c', c''$ denote the captions associated with the pre-training data. 
Additionally, $o', o''$ represent the raw image observations from the pre-training dataset, and $h_{\xi^{*}}, h', h''$ indicate the model output values.
Similar terms are added in Equations (\ref{eq:loss for phi})-(\ref{eq:loss for theta}) as well.
The overall MHCG algorithm is shown in Algorithm \ref{al:MHCG}.

\begin{algorithm}[bt]
\caption{Metropolis-Hastings Captioning Game (MHCG)}
\label{al:MHCG}
\begin{algorithmic}[1]
\State Set pre-trained network parameters $\xi^A$, $\xi^B$, $\phi^A$, $\phi^B$, $\psi^A$, $\psi^B$, $\theta^A$, $\theta^B$
\For{$r = 1$ to $R$}
    \State // \textbf{Perception} by both agents
    \For{$d = 1$ to $D$}
        \State $z^A_d \sim q(z^A_d \mid o^A_d; \psi^A)$
        \State $z^B_d \sim q(z^B_d \mid o^B_d; \psi^B)$
    \EndFor
    \State Set Sp $\gets$ A, Li $\gets$ B
    \For{$k = 1$ to $2$} 
        \For{$d = 1$ to $D$}
            \State // \textbf{Proposal} by speaker agent
            \State $c_d^{\star} \sim q(c_d^{Sp} \mid z_d^{Sp}; \xi^{Sp})$
            \State // \textbf{Judgment} by listener agent
            \State $r = \min\left(1, \frac{p(z_d^{Li} \mid c_d^{\star}; \phi^{Li})}{p(z_d^{Li} \mid c_d^{Li}; \phi^{Li})}\right)$
            \State $u \sim \text{Unif}(0, 1)$
            \If{$u \leq r$}
                \State $c_d^{Li} \gets c_d^{\star}$
            \Else
                \State $c_d^{Li} \gets c_d^{Li}$
            \EndIf
        \EndFor
        \State Swap(Sp, Li)
    \EndFor
    \State // \textbf{Learning} of text encoder and decoder
    \State $(\xi^{*}, \phi^{*}) \gets \text{Learning with Eqs. \ref{eq:loss for xi} and \ref{eq:loss for phi}}$
    \For{$d = 1$ to $D$}
        \State $z^A_d \sim p(z^A_d \mid c^A_d; \phi^A)$
        \State $z^B_d \sim p(z^B_d \mid c^B_d; \phi^B)$
    \EndFor
    \State // \textbf{Learning} of image encoder and decoder
    \State $(\psi^*, \theta^*) \gets \text{Learning with Eqs. \ref{eq:loss for psi} and \ref{eq:loss for theta}}$
\EndFor
\end{algorithmic}
\end{algorithm}

\subsection{Inter-ProbVLM}\label{sec:inter-probvlm}

In the previous MH game (i.e., MH``N''G), each agent is represented as a probabilistic generative model (PGM).
To enable the interaction between two PGMs, the previous work defined \emph{Inter-PGM}, where both PGMs share a common prior distribution.
In this study (i.e., MH``C''G), we utilize a probabilistically modeled vision-language model (VLM) to fit the MH game to variable-length discrete sequences (i.e., natural language captions).

% Similar to the structure of the previous Inter-PGM, Inter-ProbVLM is constructed by combining two probabilistic VLMs, referred to as ProbVLMs \citep{upadhyay2023probvlm}. 
Similarly to the previous Inter-PGM structures, we construct a probabilistic model we dub \emph{Inter-ProbVLM}, a combination of two probabilistic VLMs, namely \emph{ProbVLMs} \citep{upadhyay2023probvlm}.
ProbVLM serves as an adapter for models like CLIP that generate deterministic embeddings by assuming a generalized Gaussian distribution over the embeddings and producing its parameters.
The generative process of Inter-ProbVLM is described in Equations (\ref{eq:inter-pvlm-gen1})–(\ref{eq:inter-pvlm-gen3}), the graphical model is illustrated in Figure~\ref{fig:graphical_model}, 
% the graphical model decomposed for inference based on Neuro-SERKET \cite{taniguchi2020neuro} is shxown in Figure~\ref{fig:decomposed_graphical_model}, 
and the corresponding parameters are summarized in Table \ref{tab:Parameters of Inter-ProbVLM}. Here, $*\in\{A,B\}$ denotes the index representing each agent.

% \begin{figure}[tb!]
%     \centering
%     \includegraphics[keepaspectratio, scale=0.3]{graphical_model.pdf}
%     \caption{Graphical model of Inter-ProbVLM}
%     \label{fig:inter_probvlm}
% \end{figure}

\begin{table}[bt] 
    \centering 
    \resizebox{\linewidth}{!}{  % Automatically fits within column width
        \begin{tabular}{p{1.5cm} p{6cm}}  % Adjust column width as needed
            \hline
            Notation  & Description  \\
            \hline
            $D$       & Total number of observations \\  
            $c_d$     & Caption (sign) of the $d$-th observation. \\  
            $z_d^*$   & Latent variable of the $d$-th observation. \\  
            $o_d^*$   & Image of the $d$-th observation. \\  
            $\xi^*$   & Parameters of the text decoder. \\  
            $\phi^*$  & Parameters of the text encoder. \\  
            $\psi^*$  & Parameters of the image encoder. \\  
            $\theta^*$ & Parameters of the image decoder. \\  
            \hline 
        \end{tabular}
    }
    \caption{Parameters of Inter-ProbVLM. The index $*\in\{A,B\}$ represents each agent.} 
    \label{tab:Parameters of Inter-ProbVLM} 
\end{table}

\begin{align}
    c_d     &\sim p(c_d) \quad  & d = 1, \ldots, D \label{eq:inter-pvlm-gen1} \\
    z^*_d   &\sim p(z_d^* \mid c_d; \phi^*) & d = 1, \ldots, D \\
    o^*_d   &\sim p(o_d^* \mid z_d^*; \theta^*) & d = 1, \ldots, D \label{eq:inter-pvlm-gen3}
\end{align}
% Here, the prior distribution $p(c_d)$ follows Uniform Monkey Typing, a power-law-based distribution known to resemble natural language characteristics \citep{conrad2004power,ueda2024lewiss}.
The conditional distributions $p(z_d^* \mid c_d; \phi^*)$ and $p(o_d^* \mid z_d; \theta^*)$ represent probability distributions parameterized by the text encoder $\phi^*$ and the image decoder $\theta^*$, respectively. 
The intractable posteriors of \(p(z_d^* \mid c_d; \phi^*)\) and \(p(o_d^* \mid z_d; \theta^*)\) are approximated using parameters \(\xi^*\) and \(\psi^*\), forming \(q(c \mid z_d^*; \xi^*)\) and \(q(z^* \mid o^*; \psi^*)\), respectively.
Moreover, in the estimation problem depicted in Figure~\ref{fig:overview_of_research}, it holds that \(\Theta^* = \{\psi^*, \xi^*\}\).
The global parameters $\xi^*, \phi^*, \psi^*, \theta^*$ are initialized with pre-trained VLM parameters.\footnote{We do not specify the concrete structure of the prior $p(c_{d})$ that is conceptually introduced in Equation (\ref{eq:inter-pvlm-gen1}), because it is neither used during training nor during inference in our approach.
If any, it could be a language model (e.g., GPT family), as it is the probability distribution of a sentence.}

\section{Experiment 1: MHCG between VLMs pre-trained with different datasets}
\label{sec:experiment1}
% 実験
In Experiment 1, we conduct a fundamental validation of MHCG using two VLM agents pre-trained on different datasets. The evaluation focuses on whether the signs shared between agents through MHCG are plausible and whether captioning performance improves for images from the dataset used for the other agent's pre-training. If MHCG functions as a probabilistic generative model, the inferred signs are expected to reflect the fused knowledge of both agents. Furthermore, if knowledge fusion is successful, the captioning performance for images within the partner agent’s dataset should improve. To isolate the impact of linguistic representation from visual representation, the image encoder $\psi^*$ and decoder $\theta^*$ are kept fixed during this experiment.

\subsection{Datasets}
% シナリオ１では異なる大規模データセットを用いて事前学習されたエージェントによるMHCGを実現し，基本的な検証を行う．
% ２体のエージェントに事前知識を与えるためのデータセットとして，それぞれConceptual Captions 3M (CC3M)\cite{sharma2018conceptual}，MSCOCO (COCO)\cite{lin2014microsoft}を採用する．
% CC3M はWebから収集された画像とAlt-textをキャプションとしており，画像に写っていないような描写や抽象的な表現が見られるという特徴がある．
% COCOは80カテゴリのオブジェクトを含む画像にアノテーションが人手で付与されており，写っているオブジェクトについて詳細に説明されていることが特徴である．
We implement MHCG using agents pre-trained on different large-scale datasets and conduct a fundamental validation.
As datasets for providing prior knowledge to the two agents, we adopt CC3M \citep{sharma2018conceptual} and COCO \citep{lin2014microsoft}, respectively.
CC3M consists of images collected from the web with alt-text as captions, characterized by descriptions that may include content not explicitly depicted in the images or abstract expressions.
COCO contains images annotated manually with 80 object categories, featuring detailed descriptions of the objects present in the images.

\subsection{Metrics}  
To evaluate the plausibility of the shared signs, we use the log likelihood $\log p(z^A, z^B | c; \phi^A, \phi^B)$. This represents the likelihood of the generated distribution in the probabilistic generative model, indicating how plausible the captions are given the latent representations $z^A_d$ and $z^B_d$ of both agents.  

For assessing captioning performance, we employ both reference-based and reference-free metrics. The reference-based metrics include BLEU \citep{papineni2002bleu}, METEOR \citep{lavie2007meteor}, and BERT-Score \citep{zhangbertscore}. The reference-free metrics include CLIP-Score \citep{hessel2021clipscore}, PAC-Score, and RefPAC-Score \citep{sarto2023positive}.  

\subsection{Agents of Comparison}  
\begin{itemize}[leftmargin=*]
    \item \textbf{Pretrain COCO, Pretrain CC3M} are only pre-trained on COCO and CC3M, respectively.
    \item \textbf{Fine-tune COCO, Fine-tune CC3M} are fine-tuned using captions generated by their counterpart agent. The fine-tuning is applied to the each agent alternately, starting from Pretrain CC3M and Pretrain COCO. It corresponds to the special case of MHCG COCO and MHCG CC3M without acceptance-rejection mechanism ($r=1$).
    \item \textbf{MHCG COCO, MHCG CC3M} are obtained by MHCG between Pretrain COCO and Pretrain CC3M using the same parameter settings as the Fine-tune COCO and CC3M.
\end{itemize}
% - \textbf{Pretrain COCO, Pretrain CC3M}: Agents pre-trained exclusively on COCO and CC3M, respectively.  
% - \textbf{Fine-tune COCO, Fine-tune CC3M}: Agents fine-tuned using captions generated by their counterpart pre-train agent. Specifically, the Fine-tune CC3M agent is obtained by fine-tuning the Pretrain CC3M agent with captions generated by the Pretrain COCO agent, and vice versa.  
% - \textbf{MHCG COCO, MHCG CC3M}: Agents obtained by applying the proposed MHCG method to the Pretrain COCO and Pretrain CC3M agents. The parameter settings are the same as in the Fine-tune method.  

\subsection{Details}  
\begin{table}[bt]  
\small
\centering  
\scalebox{1}{  
\begin{tabular}{ll}  
\toprule  
\textbf{Parameter} & \textbf{Value} \\  
\midrule  
\multicolumn{2}{l}{\textbf{Common Settings}} \\  
MHCG epochs & 30 \\  
DER++ ($\alpha$, $\beta$) & (0.05, 0.05) \\  
LoRA ($r$, $\alpha$, dropout) & (8, 16, 0.1) \\  
\midrule  
\multicolumn{2}{l}{\textbf{Training Settings}} \\  
% Optimizer for $\xi^*$ & AdamW\cite{loshchilov2019decoupled} \\  
% Optimizer for $\phi^*$ & Adam\cite{kingma2015adam} \\  
Learning rate ($\xi^*$ / $\phi^*$) & ($1\text{e-4}$ / $1\text{e-6}$) \\  
Number of epochs & 10 \\  
Batch size & 40 \\  
\bottomrule  
\end{tabular}  
}  
\caption{Experimental settings.}  
\label{tab:experimental setting}  
\end{table}
For the pre-training of each agent's text encoder $\phi^*$ and image encoder $\psi^*$, we adopt the pre-training framework of ProbVLM \citep{upadhyay2023probvlm}. For the text decoder $\theta^*$, we employ the pre-training framework of ClipCap \citep{mokady2021clipcap}.
Table \ref{tab:experimental setting} summarizes the parameter settings adopted in our experiments.

\subsection{Results}

%\paragraph{Shared Signs}  
\paragraph{Joint Likelihood Improvement by Shared Captions}  

\begin{figure}[tb]  
    \centering  
    \includegraphics[width=0.40\textwidth]{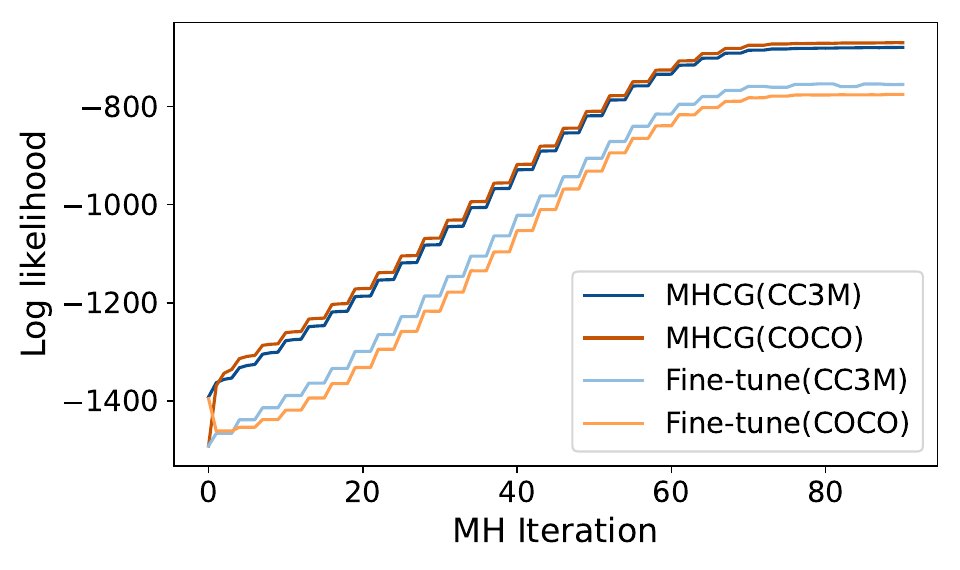} % Image filename  
    \caption{
    % Log likelihood of caption $\log p(z^A, z^B | c; \phi^A, \phi^B)$.  
    % With more MHCG iterations, the likelihood of the captions held by each agent also increases for both agents.  
    The log likelihood of a caption, $\log p(z^A, z^B \mid c; \phi^A, \phi^B)$, increases for both agents as MHCG iterations progress.
    } % Caption  
    \label{fig:likelihood of sign} % Label for reference  
\end{figure}  

Figure~\ref{fig:likelihood of sign} illustrates the transition of the likelihood $\log p(z^A, z^B | c; \phi^A, \phi^B)$ of the captions inferred by each agent, as evaluated by both agents. 
% ここでCPCに言及して「なるほど！」となる読者はほとんどいないと思う…（上田）
% Following the CPC hypothesis~\cite{taniguchi2024collective}, we assume that properly shared captions will contribute to the prediction of observation features.
As MHCG progresses, the likelihood of the captions generated by the MHCG agents improves with each iteration, surpassing that of the Fine-tune CC3M and Fine-tune COCO agents.
% Additionally, when an agent updates its parameters, a significant increase in likelihood is observed, resulting in a step-like pattern.  
% This indicates that the captions inferred through the MHCG process are shared as plausible representations for both agents, despite being trained on different datasets.  
Interestingly, at the beginning of the iterations, the likelihood of MHCG COCO increases to the same level as MHCG CC3M while the likelihood of Fine-tune COCO decreases to the same level as Fine-tune CC3M. 
This difference suggests that acceptance-rejection mechanism of MHCG contributes to prevent the agents from performance degradation in the game.

% \begin{table}[bt!p]  
%     \caption{Similarity of shared signs between agents. MHCG exhibits the highest similarity across all metrics, including BLEU@4 and METEOR, which measure word sequence similarity, as well as BERT-S, which evaluates semantic similarity.}  
%     \centering  
%     \begin{tabular}{lccc}  
%         \toprule  
%         Models & BLEU@4 & METEOR & BERT-S \\  
%         \midrule  
%         Pretrain & 3.03 & 15.28 & 0.888 \\  
%         Fine-tune & 7.36 & 23.53 & 0.906 \\  
%         MHCG & \textbf{33.93} & \textbf{49.11} & \textbf{0.934} \\  
%         \bottomrule  
%     \end{tabular}  
%     \label{tab:similarity of signs}  
% \end{table}  

% Table \ref{tab:similarity of signs} presents the similarity of the signs inferred by the two agents for the observed images through MHCG.  
% MHCG outperforms both Pretrain and Fine-tune across all similarity metrics.  
% This result not only demonstrates that both agents have learned to generate similar signs but also suggests that they have been trained to judge those signs as plausible for both agents.  

%\subsubsection{Captioning Performance}  
\paragraph{Cross-Dataset Captioning Performance}  

\begin{table*}[bt]  
\centering  
\renewcommand{\arraystretch}{0.8} % Adjust row spacing  
\resizebox{\textwidth}{!}{ % Resize to fit the page width  
\fontsize{8pt}{9.6pt}\selectfont
\begin{tabular}{clcccccc}
\toprule  
\multicolumn{2}{c}{\textbf{Agent}} & \multicolumn{6}{c}{\textbf{Validation Data of Cross-Dataset }} \\  
\cmidrule(lr){1-2} \cmidrule(lr){3-8}  
\multirow{2}{*}{Pretrain Data} & \multirow{2}{*}{Method} & \multicolumn{3}{c}{Reference-based} & \multicolumn{2}{c}{Reference-free} & Hybrid \\  
\cmidrule(lr){3-5} \cmidrule(lr){6-7} \cmidrule(lr){8-8}  
& & BLEU@4  & METEOR  & BERT-S & CLIP-S & PAC-S & RPAC-S \\  
\midrule \midrule  
\multirow{3}{*}{CC3M} & Pretrain & 6.32 & 12.93 & 0.886 & 0.747 & 0.598 & 0.695 \\  
 & Fine-tune & \textbf{26.52} & \textbf{24.39} & \textbf{0.911} & \underline{0.772} & \underline{0.619} & \underline{0.723} \\  
 & \textbf{MHCG} & \underline{17.50} & \underline{19.19} & \underline{0.901} & \textbf{0.783} & \textbf{0.625} & \textbf{0.726} \\  
\midrule \midrule  
\multirow{3}{*}{COCO} & Pretrain & 1.30 & 6.61 & 0.868 & \underline{0.720} & \underline{0.575} & \underline{0.631} \\  
 & Fine-tune & \textbf{4.19} & \textbf{8.22} & \textbf{0.875} & 0.705 & 0.563 & 0.644 \\  
 & \textbf{MHCG} & \underline{2.30} & \underline{7.74} & \underline{0.874} & \textbf{0.735} & \textbf{0.588} & \textbf{0.660} \\  
\bottomrule  
\end{tabular}  
}  
\caption{Captioning performance on the cross-dataset. The cross-dataset refers to each agent's original pre-training dataset. The highest value is bolded and the second highest is underlined.}  
\label{tab:captioning performance for cross Validation Data}  
\end{table*}  

% Table \ref{tab:captioning performance for cross Validation Data} presents the captioning performance on cross-dataset validation data, which evaluates the generalization ability of an agent to the pre-training knowledge of its counterpart.
% Here, the term ``cross-dataset'' refers to the dataset used for pre-training by the counterpart agent.
Table \ref{tab:captioning performance for cross Validation Data} presents the captioning performance on cross-dataset validation data, designed to assess how well each agent generalizes to the knowledge learned by the other agent during pre-training.
Here, “cross-dataset” refers to the dataset used in the counterpart agent’s pre-training.

The method Pretrain fails to generalize to the counterpart dataset, exhibiting consistently low performance.
The method Fine-tune achieves the highest scores in reference-based metrics, indicating that it adapts to the language expressions of cross-dataset annotations by aligning with the captions generated by the counterpart agent.
The method MHCG outperforms all methods in reference-free and hybrid metrics.
This result suggests that, through the judgment of acceptance-rejection based on the acceptance probability in Equation (\ref{eq:acceptance probability}), MHCG selects the most semantically plausible captions for the given images, enabling effective parameter updates.
Furthermore, MHCG surpasses Pretrain across all metrics, demonstrating that the agents successfully acquire knowledge from their counterparts.

\paragraph{Original Dataset Captioning Performance}  

\begin{table*}[bt]  
\centering  
\renewcommand{\arraystretch}{0.8} % Adjust row spacing  
\resizebox{\textwidth}{!}{ % Resize to fit the page width  
\fontsize{8pt}{9.6pt}\selectfont
\begin{tabular}{clcccccc}  
\toprule  
\multicolumn{2}{c}{\textbf{Agent}} & \multicolumn{6}{c}{\textbf{Validation Data of Original-Dataset}} \\  
\cmidrule(lr){1-2} \cmidrule(lr){3-8}  
\multirow{2}{*}{Pretrain Data} & \multirow{2}{*}{Method} & \multicolumn{3}{c}{Reference-based} & \multicolumn{2}{c}{Reference-free} & Hybrid \\  
\cmidrule(lr){3-5} \cmidrule(lr){6-7} \cmidrule(lr){8-8}  
& & BLEU@4  & METEOR  & BERT-S & CLIP-S & PAC-S & RPAC-S \\  
\midrule \midrule  
\multirow{3}{*}{CC3M} & Pretrain & \textbf{7.42} & \textbf{11.14} & \textbf{0.881} & \underline{0.765} & \underline{0.612} & \textbf{0.686} \\  
 & Fine-tune & 1.64 & 7.16 & 0.868 & 0.717 & 0.574 & 0.635 \\  
 & \textbf{MHCG} & \underline{3.50} & \underline{9.09} & \underline{0.877} & \textbf{0.768} & \textbf{0.614} & \underline{0.683} \\  
\midrule \midrule  
\multirow{3}{*}{COCO} & Pretrain & \textbf{31.61} & \textbf{27.37} & \textbf{0.918} & \textbf{0.808} & \textbf{0.645} & \textbf{0.750} \\  
 & Fine-tune & 9.59 & 15.00 & 0.892 & 0.747 & 0.599 & 0.703 \\  
 & \textbf{MHCG} & \underline{20.77} & \underline{20.72} & \underline{0.904} & \underline{0.788} & \underline{0.629} & \underline{0.732} \\  
\bottomrule  
\end{tabular}  
}  
\caption{Captioning performance on the original-dataset. The original dataset refers to the dataset used for each agent's pre-training.}  
\label{tab:captioning performance for original Validation Data}  
\end{table*}  

% Table \ref{tab:captioning performance for original Validation Data} presents the captioning performance on the validation data, evaluating whether the agents can retain their pre-training knowledge.  
% Here, the original dataset refers to the dataset used for pre-training the agent, and this evaluation examines whether the pre-training knowledge is preserved.
Table \ref{tab:captioning performance for original Validation Data} shows the captioning performance on each agent’s original pre-training dataset, evaluating how well the agents retain their pre-trained knowledge.
The Pretrain agents generalize relatively well to their own dataset, achieving the highest scores across almost all metrics.  
This result suggests that forgetting occurs in both the Fine-tune and MHCG models.  
Fine-tune directly learns from captions generated by the counterpart agent, leading to over-adaptation to those captions and, consequently, greater forgetting of its own pre-trained knowledge.  
In contrast, while some forgetting is also observed in MHCG, its extent is lower than that of Fine-tune.  
% This can be attributed to MHCG's judgment process, where captions that significantly deviate from its own knowledge are rejected, allowing the agent to retain more of its original knowledge.
This indicates the effect of MHCG’s judgment process, which rejects captions deviating from the agent’s own knowledge and helping preserve its original understanding.

\paragraph{Examples of Generated Captions}
Table \ref{tab:generated_captions_sample} presents examples of captions generated by the agents.  
The Pretrain COCO agent incorrectly describes the content as a ``white bird.''  
In contrast, the MHCG COCO agent correctly generates ``honeybee'' and ``vector illustration.''  
Notably, the terms ``vector illustration'' and ``honeybee'' exist in the CC3M pre-training data but are absent from the COCO pre-training data.  
This indicates that, through MHCG, the COCO agent acquired knowledge from the CC3M agent’s captions without requiring direct access to the CC3M dataset.  
Additionally, MHCG COCO shows the capability of describing text appearing in the image, such as ``name and address.''

\begin{table*}[bt]  
  \small
  \centering 
  \renewcommand{\arraystretch}{0.7} % Adjust row spacing  
  \begin{tabular}{p{1.7cm}clp{9.8cm}} % Image on the left, caption text on the right  
    \toprule  
    \multirow{2}{*}{\textbf{Input image}} & \multicolumn{2}{c}{\textbf{Agent}} & \multirow{2}{*}{\textbf{Caption}} \\   
    \cmidrule(lr){2-3}  
    & Pretrain & Method & \\  
    \midrule  
    \multirow{6}{*}{\scalebox{0.48}{\includegraphics{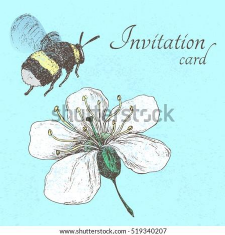}}} & \multirow{3}{*}{CC3M} & Pretrain & honeybee with a flower in the beehive. \\  
     & & Fine-tune & a picture of flowers in a vase with some bees. \\  
     & & \textbf{MHCG} & \textbf{a painting of a bee with flowers.} \\  
     \cmidrule{2-4}  
     & \multirow{3}{*}{COCO} & Pretrain & A picture of a white bird with flowers on it. \\  
     & & Fine-tune & vector illustration of a honeybee. \\  
     & & \textbf{MHCG} & \textbf{vector illustration of a honeybee with the name and address.} \\
    \bottomrule  
  \end{tabular}  
  \caption{Examples of captions generated by each agent for input images.} 
  \label{tab:generated_captions_sample} % Table label  
\end{table*} 

\section{Experiment 2: Fine-grained Analysis of MHCG Effect through Category-level Comparison}
\label{sec:experiment2}
In Experiment 2, we investigate more detailed evaluation of the effectiveness of MHCG.
We partition the COCO dataset based on categories into subsets thereby explicitly separating the vocabulary that each agent can learn during pre-training.  
This enables a comprehensive comparison with existing methods across the entire dataset.

\subsection{Splitting COCO Dataset}
We partition the original COCO dataset into three subsets, $\mathrm{COCO\text{-}a}$, $\mathrm{COCO\text{-}b}$, and $\mathrm{COCO_{Other}}$, based on the annotation for each image.

First, we add a predefined common category to both $\mathrm{COCO\text{-}a}$ and $\mathrm{COCO\text{-}b}$. Next, for each super-category in COCO, we arrange the remaining categories—after excluding the common category—in descending order according to the number of images. Within each super-category, we sequentially assign one category at a time to the dataset that currently has fewer images or fewer categories, thereby ensuring a balanced distribution between the two subsets. This assignment process takes into account not only the number of categories but also the number of images contained in each category, resulting in a more uniform data distribution.

An image is assigned to $\mathrm{COCO\text{-}a}$ (resp. $\mathrm{COCO\text{-}b}$) if all the categories annotated with the image are included in those of $\mathrm{COCO\text{-}a}$ (resp. $\mathrm{COCO\text{-}b}$).
Otherwise, it is assigned to $\mathrm{COCO_{Others}}$.
% Image assignment is performed such that an image is allocated to a dataset if all of its annotated categories are entirely contained within the category set of either $\mathrm{COCO\text{-}a}$ or $\mathrm{COCO\text{-}b}$. Otherwise, the image is managed separately as part of $\mathrm{COCO_{Others}}$. This approach guarantees that images based on exclusive categories are allocated in a mutually exclusive manner within each subset and provides a flexible framework that can be extended to three or more partitions.

We adopt ``person'' as the predefined common category.
Consequently, both datasets retain ``person'' as a shared common element, while all other categories are exclusively assigned to either $\mathrm{COCO\text{-}a}$ or $\mathrm{COCO\text{-}b}$. For training, the partitioned $\mathrm{COCO\text{-}a}$ and $\mathrm{COCO\text{-}b}$ are used as the pre-training datasets, and an additional 15,000 images extracted from those not employed in pre-training—designated as $\mathrm{COCO_{Others}}$—are used as the dataset for MHCG.

\subsection{Metrics for Category-level Comparison}
% In Experiment 2, we construct an evaluation metric to quantify category-level comparison by determining whether synonyms of a category appear in the generated captions. 
% Here, category-level comparison refers to the ability, in the image captioning task, to generate words or phrases corresponding to the annotated categories in an image. 
The category-level comparison is also hold in the image captioning task.
We construct evaluation metrics based on finding the synonyms of the COCO categories described above in the generated caption.

Specifically, similar to the evaluation metrics for multi-label classification \citep{liu2021query2label,luo2019visual}, we introduce the following overall evaluation (Overall) and category-level evaluation (Per-Category) metrics:
\begin{align} 
\mathrm{OP}^S &= \frac{\sum_i M_c^i}{\sum_i M_p^i}, \quad 
\mathrm{OR}^S = \frac{\sum_i M_c^i}{\sum_i M_g^i}, \notag \\ 
\mathrm{CP}^S &= \frac{1}{C} \sum_i \frac{M_c^i}{M_p^i}, \quad 
\mathrm{CR}^S = \frac{1}{C} \sum_i \frac{M_c^i}{M_g^i}, \notag \\ 
\mathrm{OF1}^S &= \frac{2 \times \mathrm{OP}^S \times \mathrm{OR}^S}{\mathrm{OP}^S + \mathrm{OR}^S}, \quad \notag\\
\mathrm{CF1}^S &= \frac{2 \times \mathrm{CP}^S \times \mathrm{CR}^S}{\mathrm{CP}^S + \mathrm{CR}^S}, 
\end{align}
where $M_c^i$ is the number of images in which synonyms related to category $i$ are correctly included, $M_p^i$ is the number of images where synonyms appear in the generated captions, and $M_g^i$ is the ground-truth number of images.
$C$ represents the total number of categories.

Synonyms for each category are extracted using cosine similarity of embedding vectors obtained from Sentence-BERT \citep{reimers2019sentence}, following the approach of \citealt{petryk2024aloha}. Specifically, noun phrases consisting of 1-gram and 2-gram terms are extracted from the COCO caption training data, and the top $K$ most similar phrases (where $K=5$ in our experiments) are selected by comparing them against other category names and supercategories. 

\subsection{Comparison Methods}  
In Experiment 2, we introduce additional comparison methods to evaluate the overall performance across both datasets. 
\begin{itemize}[leftmargin=*]
    \item \textbf{Pretrain $\mathbf{COCO\text{-}all}$}: An agent pre-trained on the entire COCO dataset. This serves as the topline in this experiment.  
    \item \textbf{Weight Averaging}: An agent obtained by merging two pre-trained agents through averaging their weights in the parameter space \citep{wortsman2022modelsoup}.
    \item \textbf{Ensemble}: A method that generates the next token by averaging the logits output from the Pretrain $\mathrm{COCO\text{-}a}$ and Pretrain $\mathrm{COCO\text{-}b}$ agents \citep{jiang2023llm}.
    \item \textbf{PackLLM}: A method that ensembles the logits outputs by applying perplexity-based weighting \citep{mavromatis2024pack}.
    \item \textbf{KD $\mathbf{COCO\text{-}a}$, KD $\mathbf{COCO\text{-}b}$}: Agents fine-tuned via Knowledge Distillation (KD) to minimize the KL divergence with the output probability distributions of their pre-trained counterparts \citep{timiryasov2023baby}.
\end{itemize}

% \red{Explanations of KD, weight averaging, packllm}
% KD: 相手のpre-train agentが出力した確率分布とのKLダイバージェンスが低くなるようにファインチューニングしたエージェント\cite{}．
% Weight Averaging: パラメータ空間で，２体のpre-train agent の重みを平均することでマージしたエージェント\cite{}．
% PackLLM: トークンのパープレキシティに基づいて重み付けして logits outputをアンサンブルするメソッド．

\begin{table*}[t]
  \centering
  \renewcommand{\arraystretch}{0.9} % Adjust line spacing
  {\fontsize{10pt}{12pt}\selectfont
    \begin{tabular}{lclp{8.9cm}}
      \toprule
      \multirow{2}{*}{\textbf{Input image}} & \multicolumn{2}{c}{\textbf{Agent}} & \multirow{2}{*}{\textbf{Caption}} \\ 
      \cmidrule(lr){2-3}
      & Pretrain & Method & \\
      \midrule
      \multirow{12}{*}{
        \begin{minipage}{0.15\textwidth}
          \includegraphics[width=\textwidth]{coco_a_example.pdf}\\
          \centering \small Image ($\mathrm{COCO\text{-}a}$)
        \end{minipage}
      }
        & $\mathrm{COCO\text{-}all}$ & Pretrain   & a couple of men standing in a field with one holding a \blue{frisbee}. \\
      \cmidrule{2-4}
        & $\mathrm{COCO\text{-}a}$    & Pretrain   & Three men are carrying \blue{Frisbees} on a field. \\
        & $\mathrm{COCO\text{-}b}$    & Pretrain   & Three people playing \red{baseball} in the grass at a baseball field. \\
      \cmidrule{2-4}
        & \multicolumn{2}{c}{Weight Averaging}   & a couple of men play a game of ping pong  in front of a \red{car}.\\
        & \multicolumn{2}{c}{Ensemble}   & a man standing next to two other men on a field with a \blue{frisbee}. \\       
        & \multicolumn{2}{c}{PackLLM}   & Two men stand on a grassy field holding \blue{frisbees}. \\
        & $\mathrm{COCO\text{-}a}$    & Fine-tune  & two men are holding up their \red{baseball bats} in the grass. \\
        & $\mathrm{COCO\text{-}b}$    & Fine-tune  & a man holding a \blue{frisbee} while another stands nearby. \\
        & $\mathrm{COCO\text{-}a}$    & KD  & A man holding a \red{bat} in the grass while two other men watch. \\
        & $\mathrm{COCO\text{-}b}$    & KD  & a group of men playing with a \blue{frisbee} on the grass. \\
        & $\mathrm{COCO\text{-}a}$    & \textbf{MHCG}    & \textbf{a group of men standing around a field holding \blue{frisbees}.} \\
        & $\mathrm{COCO\text{-}b}$    & \textbf{MHCG}    & \textbf{a group of men walking down a field with \blue{frisbees}.} \\
      \bottomrule
    \end{tabular}
  }
  \vspace{-2ex}
  \caption{Examples of captions generated by each agent for the image in $\mathrm{COCO\text{-}a}$ dataset. 
  \blue{Blue} denotes categories present in the image, while \red{red} indicates those absent.
  % 青は画像に含まれるカテゴリの単語、赤は画像内に含まれないカテゴリの単語である。
  }
  \vspace{-2ex}
  \label{tab:generated_captions_sample_split_coco}
\end{table*}

\begin{table*}[!h]
\centering
\renewcommand{\arraystretch}{0.8} % Adjust line spacing
\resizebox{\textwidth}{!}{ % Resize to fit the page width
\fontsize{8pt}{9.6pt}\selectfont
\begin{tabular}{clccccccc}
\toprule
\multicolumn{2}{c}{\textbf{Agent}} & \multicolumn{7}{c}{\textbf{Overall Dataset}} \\
\cmidrule(lr){1-2} \cmidrule(lr){3-9} 
Pretrain Data & Method & $\mathrm{OP}^S \uparrow$ & $\mathrm{OR}^S \uparrow$ & $\mathrm{OF1}^S \uparrow$ & $\mathrm{CP}^S \uparrow$ & $\mathrm{CR}^S \uparrow$ & $\mathrm{CF1}^S \uparrow$ & Time $\downarrow$ \\
\midrule \midrule
\rowcolor[gray]{0.96}
Topline &  &  &  &  &  &  &  &  \\
$\mathrm{COCO\text{-}all}$ & Pretrain  & 0.758 & 0.351 & 0.480 & 0.733 & 0.416 & 0.530 & 1.00 \\
\midrule
\rowcolor[gray]{0.96}
Base Agents &  &  &  &  &  &  &  &  \\
$\mathrm{COCO\text{-}a}$ & Pretrain & 0.587 & 0.235 & 0.335 & 0.587 & 0.272 & 0.372 & 1.00 \\
$\mathrm{COCO\text{-}b}$ & Pretrain & 0.563 & 0.230 & 0.326 & 0.538 & 0.263 & 0.354 & 1.00 \\
\midrule
\rowcolor[gray]{0.96}
\multicolumn{2}{l}{Fusion Methods} &  &  &  &  &  &  &  \\
\multicolumn{2}{c}{Weight Averaging} & 0.035 & 0.006 & 0.010 & 0.028 & 0.005 & 0.009 & \textbf{1.00} \\
\multicolumn{2}{c}{Ensemble} & 0.664 & 0.255 & 0.368 & \underline{0.663} & 0.295 & 0.408 & 1.85 \\
\multicolumn{2}{c}{PackLLM} & 0.605 & 0.241 & 0.345 & 0.587 & 0.278 & 0.377 & 2.00 \\
$\mathrm{COCO\text{-}a}$ & Fine-tune & 0.578 & 0.248 & 0.347 & 0.601 & 0.282 & 0.384 & \textbf{1.00} \\
$\mathrm{COCO\text{-}b}$  & Fine-tune & 0.575 & 0.249 & 0.347 & 0.581 & 0.290 & 0.387 & \textbf{1.00} \\
$\mathrm{COCO\text{-}a}$ & KD  & 0.557 & 0.227 & 0.322 & 0.564 & 0.259 & 0.355 & \textbf{1.00} \\
$\mathrm{COCO\text{-}b}$ & KD  & 0.546 & 0.228 & 0.322 & 0.547 & 0.267 & 0.358 & \textbf{1.00} \\
 $\mathrm{COCO\text{-}a}$& \textbf{MHCG} & \textbf{0.715} & \textbf{0.281} & \textbf{0.404} & \textbf{0.670} & \textbf{0.345} & \textbf{0.455} & \textbf{1.00} \\
$\mathrm{COCO\text{-}b}$ & \textbf{MHCG} & \underline{0.689} & \underline{0.276} & \underline{0.394} & 0.645 & \underline{0.338} & \underline{0.444} & \textbf{1.00} \\
\bottomrule
\end{tabular}
}

\caption{Metrics of category-level comparison and generation time over the overall COCO dataset.  
}
\vspace{-2ex}

\label{tab:category_base_metrics_on_overall-dataset}
\end{table*}

\begin{figure*}[!t]
    \centering
    \includegraphics[width=\textwidth]{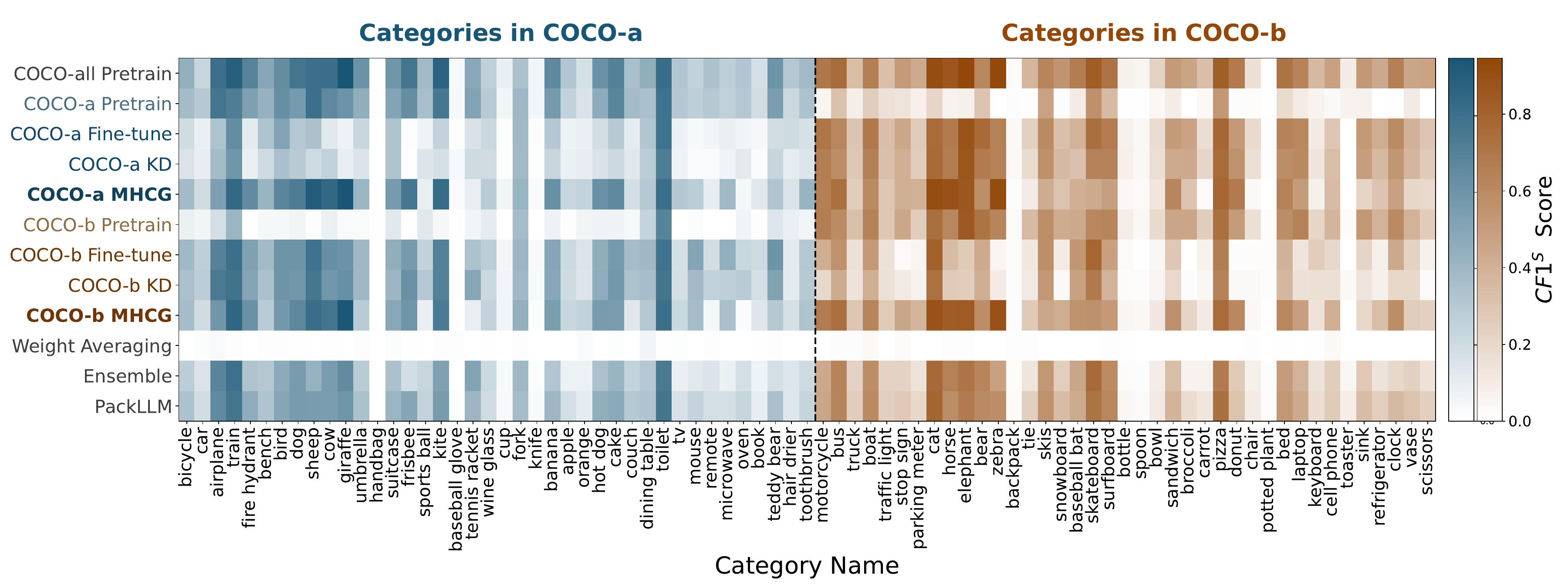} % image file name
  \vspace{-2ex}
    \caption{Heat map of the category-level comparison of F1 scores, $\mathrm{CF1}^S$.}
    \label{fig:heat_map_cf1_scores} % reference label
  \vspace{-1ex}
\end{figure*}

\subsection{Results}

\paragraph{Examples of Generated Captions}  
Table \ref{tab:generated_captions_sample_split_coco} shows examples of captions generated for an image from the $\mathrm{COCO\text{-}a}$ dataset. At the pre-training stage, the $\mathrm{COCO\text{-}a}$ agent, equipped with prior knowledge, correctly generates ``Frisbees,'' whereas the $\mathrm{COCO\text{-}b}$ agent, lacking information about the target category, erroneously generates ``baseball.'' During fine-tuning, the $\mathrm{COCO\text{-}a}$ agent is influenced by the misgenerated output from $\mathrm{COCO\text{-}b}$ and generates ``baseball bats,'' while the $\mathrm{COCO\text{-}b}$ agent, having acquired complementary information, outputs ``frisbee.'' In contrast, when the proposed MHCG is applied, both agents effectively mitigate the forgetting of their own knowledge while successfully acquiring semantic information from the counterpart, resulting in the accurate generation of ``frisbees.''

\paragraph{Overall Category-level Comparison}
Table \ref{tab:category_base_metrics_on_overall-dataset} summarizes the category-level comparison of the overall COCO dataset and caption generation time.
The Topline $\mathrm{COCO\text{-}all}$ agent exhibits the highest scores across all metrics.  
Fine-tune and KD agents perform comparably or slightly better than Pretrain agents, due to gains in counterpart categories but with some forgetting of their own.
Weight averaging cannot accommodate scenarios with different pre-training settings, resulting in fusion failure and lower scores.
In contrast, MHCG agents outperform comparison methods across almost all measures.  
Notably, the $\mathrm{COCO\text{-}a}$ MHCG agent outperforms both Ensemble and PackLLM, which require 1.85 and 2.00 times longer generation times, respectively.
Thus, MHCG agents effectively generate captions with enhanced category-level comparison in image captioning task.

\paragraph{Visualization of Category-level Comparison}
Figure~\ref{fig:heat_map_cf1_scores} presents a heat map of the $\mathrm{CF1}^S$ scores for each category. 
The left side displays categories originally belonging to $\mathrm{COCO\text{-}a}$, while the right side corresponds to those from $\mathrm{COCO\text{-}b}$.

Pretrain agents demonstrate high $\mathrm{CF1}^S$ for categories encountered during pre-training and low $\mathrm{CF1}^S$ for unseen ones.  
For example, the $\mathrm{COCO\text{-}a}$ Pretrain agent shows high scores for $\mathrm{COCO\text{-}a}$ categories and uniformly low scores for $\mathrm{COCO\text{-}b}$ categories.  
Conversely, the $\mathrm{COCO\text{-}b}$ Pretrain agent exhibits the opposite pattern.  

Fine-tune and KD agents suffer from significant forgetting of their originally learned categories.  
The $\mathrm{COCO\text{-}a}$ Fine-tune agent, for instance, shows reduced $\mathrm{CF1}^S$ for $\mathrm{COCO\text{-}a}$ categories while exhibiting improved scores for $\mathrm{COCO\text{-}b}$ categories.  
Similarly, the $\mathrm{COCO\text{-}b}$ Fine-tune agent shows high scores for $\mathrm{COCO\text{-}a}$ and low scores for $\mathrm{COCO\text{-}b}$.  
These trends are also observed in KD.
This pattern suggests that fine-tuning with captions (or probability in KD) generated by the opposite agent leads to the acquisition of their knowledge at the cost of forgetting one's own.  

In contrast, the MHCG agents achieve a balance by preserving their inherent knowledge while enhancing $\mathrm{CF1}^S$ of the counterpart categories.  
Although the $\mathrm{COCO\text{-}a}$ MHCG agent shows slight forgetting in some categories (e.g., \textit{tennis racket}), it largely retains the $\mathrm{CF1}^S$ levels of its original categories.  
Moreover, the $\mathrm{CF1}^S$ of $\mathrm{COCO\text{-}b}$ categories improves similar to the Fine-tune case.  
Compared to ensemble-based methods, MHCG demonstrates superior $\mathrm{CF1}^S$ across many categories—from larger objects such as "giraffe" and "zebra" to smaller items like "clock" and "kite."
As a result, the MHCG agents produce performance that closely resemble those of the Topline $\mathrm{COCO\text{-}all}$ agent.

\section{Conclusion}
In this paper, we proposed MHCG, a probabilistic framework in which VLM agents with different prior knowledge generate mutually plausible captions and learn from them to fuse their knowledge.
Experiments on VLM agents pre-trained with CC3M and COCO datasets showed that as MHCG iterations increases, the knowledge, namely domain-specific vocabulary, becomes more plausible for both agents, leading to improved captioning performance on the counterpart's pre-training data.
This improvement is attributed to the acceptance-rejection process based on the Metropolis-Hastings algorithm, which ensures that agents learn from more plausible captions, reinforcing knowledge fusion while mitigating catastrophic forgetting.
Furthermore, experiments on the split COCO dataset demonstrated that MHCG achieved higher score in category-level comparison than existing approaches without increasing captioning time.

%谷口のidea
MHCG has demonstrated effectiveness in fusing knowledge between two VLMs, but further extensions are necessary. First, scaling to multiple agents could enhance decentralized knowledge fusion~\cite{inukai2023recursive}. Second, integrating VLMs trained in different languages or with imbalanced datasets remains a challenge. Future work should explore MHCG’s adaptability across diverse scenarios.

\section*{Acknowledgements}
This study was partially supported by the Japan Society for the Promotion of Science (JSPS) KAKENHI under Grants JP21H04904, and JP23H04835.

\bibliography{tacl2021}
\bibliographystyle{acl_natbib}

\iftaclpubformat

\onecolumn

% \appendix
% \section{Author/Affiliation Options as set forth by MIT Press}
% \label{sec:authorformatting}

% Option 1. Author’s address is underneath each name, centered.

% \begin{quote}\centering
%   \begin{tabular}{c}
%     \textbf{First Author} \\
%     First Affiliation \\
%     First Address 1 \\
%     First Address 2 \\
%     \texttt{first.email@example.com}
%   \end{tabular}
%   \ 
%   \begin{tabular}{c}
%     \textbf{Second Author} \\
%     Second Affiliation \\
%     Second Address 1 \\
%     Second Address 2 \\
%     \texttt{second.email@example.com}
%   \end{tabular}

%   \begin{tabular}{c}
%     \textbf{Third Author} \\
%     Third Affiliation \\
%     Third Address 1 \\
%     Third Address 2 \\
%     \texttt{third.email@example.com}
%   \end{tabular}
% \end{quote}

% Option 2. Author’s address is linked with superscript characters to its name,
% author names are grouped, centered.

% \begin{quote}\centering
%     \textbf{First Author$^\diamond$} \quad \textbf{Second Author$^\dagger$} \quad
%     \textbf{Third Author$^\ddagger$}
%     \\ \ \\
%     $^\diamond$First Affiliation \\
%     First Address 1 \\
%     First Address 2 \\
%     \texttt{first.email@example.com}
%      \\ \ \\
%      $^\dagger$Second Affiliation \\
%     Second Address 1 \\
%     Second Address 2 \\
%     \texttt{second.email@example.com}
%      \\ \ \\
%     $^\ddagger$Third Affiliation \\
%     Third Address 1 \\
%     Third Address 2 \\
%     \texttt{third.email@example.com}
% \end{quote}
  
% \fi

\end{document}